# Object Detection Using Sim2Real Domain Randomization for Robotic Applications

Dániel Horváth , Gábor Erdős , Zoltán Istenes , Tomáš Horváth , and Sándor Földi

*Abstract*—Robots working in unstructured environments must be capable of sensing and interpreting their surroundings. One of the main obstacles of deep-learning-based models in the field of robotics is the lack of domain-specific labeled data for different industrial applications. In this article, we propose a sim2real transfer learning method based on domain randomization for object detection with which labeled synthetic datasets of arbitrary size and object types can be automatically generated. Subsequently, a state-of-the-art convolutional neural network, YOLOv4, is trained to detect the different types of industrial objects. With the proposed domain randomization method, we could shrink the reality gap to a satisfactory level, achieving 86.32% and 97.38% $\mathrm{mAP}_{50}$ scores, respectively, in the case of zero-shot and one-shot transfers, on our manually annotated dataset containing 190 real images. Our solution fits for industrial use as the data generation process takes less than 0.5 s per image and the training lasts only around 12 h, on a GeForce RTX 2080 Ti GPU. Furthermore, it can reliably differentiate similar classes of objects by having access to only one real image for training. To our best knowledge, this is the only work thus far satisfying these constraints.

Manuscript received 26 January 2022; revised 20 June 2022; accepted 24 August 2022. This work was supported in part by "Research on prime exploitation of the potential provided by the industrial digitalisation," under Grant ED_18-2-2018-0006 and in part by the Ministry for Innovation and Technology and the National Research, Development and Innovation Office within the framework of the National Lab for Autonomous Systems. The work of Tomáš Horváth was supported in part by the project "Application Domain Specific Highly Reliable IT Solutions" financed by the National Research, Development and Innovation Fund of Hungary under Grant TKP2020-NKA-06, and in part by the Thematic Excellence Programme under Grant 2020-4.1.1.-TKP2020 (National Challenges Subprogramme) funding scheme. This paper was recommended for publication by Associate Editor J. Civera and Editor W. Burgard upon evaluation of the reviewers' comments. *(Corresponding author: Dániel Horváth.)*

Dániel Horváth is with the Centre of Excellence in Production Informatics and Control, Institute for Computer Science and Control, Eötvös Loránd Research Network, 1111 Budapest, Hungary, and also with the CoLocation Center for Academic and Industrial Cooperation, Eötvös Loránd University, 1117 Budapest, Hungary (e-mail: daniel.horvath@sztaki.hu).

Gábor Erdős is with the Centre of Excellence in Production Informatics and Control, Institute for Computer Science and Control, Eötvös Loránd Research Network, 1111 Budapest, Hungary, and also with the Department of Manufacturing Science and Engineering, Budapest University of Technology and Economics, 1111 Budapest, Hungary (e-mail: erdos.gabor@sztaki.hu).

Zoltán Istenes is with the CoLocation Center for Academic and Industrial Cooperation, Eötvös Loránd University, 1117 Budapest, Hungary (e-mail: istenes@inf.elte.hu).

Tomáš Horváth is with the Department of Data Science and Engineering, Eötvös Loránd University, 1117 Budapest, Hungary, and also with the Institute of Computer Science, Faculty of Science, Pavol Jozef Šafárik University, 040 01 Košice, Slovakia (e-mail: tomas.horvath@inf.elte.hu).

Sándor Földi is with the Centre of Excellence in Production Informatics and Control, Institute for Computer Science and Control, Eötvös Loránd Research Network, 1111 Budapest, Hungary (e-mail: sandorfoldi98@gmail.com).

Color versions of one or more figures in this article are available at https://doi.org/10.1109/TRO.2022.3207619.

Digital Object Identifier 10.1109/TRO.2022.3207619

*Index Terms*—Computer vision for automation, deep learning in robotics and automation, localization, sim2real knowledge transfer.

## I. INTRODUCTION

NEW-GENERATION intelligent manufacturing (NGIM) is a recent trend embodying the in-depth integration of new-generation artificial intelligence (AI) with advanced manufacturing technology such as robotics. It became the driving force of the fourth industrial revolution and it belongs to the human–cyber-physical system 2.0 (HCPS 2.0) framework. Contrary to traditional manufacturing where robots work in structured environments and perform high-accuracy, repetitive tasks with minimal sensory input, an NGIM system is designed to be flexible and to take over as much intellectual and manual labor as possible. Thus, human workforce can concentrate on more valuable creative work. [1]

Consequently, the main interest has been shifting toward adaptive robotic applications that can cost-efficiently handle low-quantity customized products and integrate human operators with different skills and abilities. Computer vision plays an essential role here—the highly researched field has already proven useful in pick-and-place, bin picking, grasping, navigation, or quality assurance tasks. As two examples, Zeng et al. [2] created a vision-based model that can predict parameters of motion primitives through trial and error for robotic grasping and throwing, and Alonso et al. [3] designed a model for real-time semantic segmentation of RGB images for a mobile robot application.

In a computer vision context, deep convolutional neural networks (DCNNs) have been performing incredibly well on large public datasets such as ImageNet [4] or MS COCO [5]. Having reached human-level performance in classification, the main focus of computer vision research has shifted to object detection and led to networks such as the faster R-CNN [6], single shot multibox detector (SSD) [7], and YOLOv4 [8]. Even though these networks are outperforming the traditional machine-vision-based methods by a significant margin, their application in robotic systems has its difficulties. One of the main obstacles in applying deep-learning-based models is that they need to be trained on a sufficiently large, domain-specific, and expertly labeled dataset.

Levine et al. [9] conducted two experiments, in order to create a dataset of real images for their DCNN model predicting the success of robotic grasp attempts, as well as to control these attempts. Yielding records of more than 1.7 million grasp attempts with the simultaneous use of 6–14 robots, the process took months to complete. This example shows that, in general, collecting a dataset from the real world not only requires an immense amount of resources but is a time-consuming process as well.

The main motivation behind transfer learning is to overcome the aforementioned obstacle by transferring knowledge between tasks or domains [10], [11]. Sim2real transfer is a special case of transfer learning, where the source domain is the virtual simulation of the real world, while the target domain is the physical reality itself [12]. With sim2real knowledge transfer, the model can be trained in a virtual simulation, having the necessary amount of labeled synthetic data. In the case of computer vision, the images are rendered, and the labels, too, can be generated for them in a self-supervised way. Thus, the time-consuming process of data collection and labeling can be omitted. As the domains of training and test datasets are inherently different, ceteris paribus, the learned model will perform poorly in the target domain. The phenomenon of performance loss from the simulation to the real domain is called *reality gap*. Domain adaptation [13] and domain randomization [14], [15] are two ways of shrinking this gap.

The contributions of this article are as follows.
1) A sim2real domain randomization method for object detection which describes a data generation process with our domain randomization methods, a model training phase, and an evaluation phase.
2) A real-world dataset of 190 manually annotated images (RGB and depth) containing 920 objects of ten classes that address the problem of high class similarity to validate our method. The dataset is publicly available alongside our code and can serve as a benchmark for object detection algorithms.
3) For evaluation, we introduced an altered type of confusion matrix fit to object detection. It proved to be extremely useful for detecting and quantifying misclassifications which is the primary cause of performance loss in the case of similar classes.
4) A real-world robotic implementation of the method as a proof of concept containing an ROS-based robot control system.
5) Our implementations of our sim2real data generation and training module, and our robot control framework that are available at[1,2]. Both can be used as out-of-the-box software modules for industrial robotic applications.

Results of the article are as follows.
1) We achieved 86.32% $mAP_{50}$ and 97.38% $mAP_{50}$ scores in zero-shot and one-shot transfers that show the usability of our methods even in the case of an industrial application where high reliability is crucial.
2) Our experiments show that having even only one sample image from the target domain significantly improves the model's performance for similar classes.
3) A thorough ablation study focusing on finding the key factors of the data generation process.

The industrial benefit of our work is a freely available tool streamlining the training of CNN-based models for object detection. Our built-in sim2real domain randomization method spares the user the effort of collecting and annotating a large dataset, as it automatically generates training data from 3-D models. Optionally, one annotated image with all relevant objects can be added for improved performance. With training being automated as well, the entire workflow from 3-D models to trained CNN takes only around 13 h.

The rest of this article is organized as follows. Sections II and III present the problem statement and related work. In Sections IV and V, our method is outlined with the evaluation metrics. In Sections VI and VII, our dataset and our results are shown. Additionally, in Section VIII, a thorough ablation study of our method is presented. Section IX shows a real-world robotic implementation of our method. Finally, Section X concludes this article.

## II. PROBLEM STATEMENT

The main problem we tackle is how to transfer knowledge efficiently from simulation to the real world in the case of object detection. In the common computer vision problem of detecting objects, the model is given an image in which it finds center points and dimensions of bounding boxes around objects and classifies the latter. Knowledge transfer, which is the primary focus of our article, belongs to the field of transfer learning. Additional challenges arise from the following circumstances.
1) Having no or only one image from the real world.
2) Having industrial objects that share similar features, thus, making it more challenging to properly classify them.

In the following, we give a brief overview of transfer learning which is the main field of this article.

Pan et al. [10] and Weiss et al. [11] define transfer learning in their surveys. A domain $\mathcal{D}$ is defined with a feature space $\mathcal{X}$ and a probability distribution $P(X)$, where $X = x_1, x_2, x_3, \ldots, x_n \in \mathcal{X}$, while a task $\mathcal{T}$ is defined with a label space $\mathcal{Y}$ and a predictive function $f(\cdot)$. From a probabilistic point of view, $f(\cdot)$ can be seen as $P(Y \mid X)$. Thus, a domain $\mathcal{D} = \{\mathcal{X}, P(X)\}$ and a task $\mathcal{T} = \{\mathcal{Y}, P(Y \mid X)\}$. Given a specific source domain and task pair $\{\mathcal{D}_S, \mathcal{T}_S\}$ and a specific target domain and task pair $\{\mathcal{D}_T, \mathcal{T}_T\}$, transfer learning can be defined as the process of increasing the performance of the target predictive function $f_T(\cdot)$ with the help of knowledge gained from $\{\mathcal{D}_S, \mathcal{T}_S\}$, where $\mathcal{D}_S \neq \mathcal{D}_T$ or $\mathcal{T}_S \neq \mathcal{T}_T$.

An example of $\mathcal{T}_S \neq \mathcal{T}_T$ is when an image classifier, trained on a large public dataset, is reused and altered to perform object detection on the same domain ($\mathcal{D}_S = \mathcal{D}_T$). In this case, the label spaces are different, and, consequently, the conditional probability distributions of the inputs and the outputs $P(Y \mid X)$ are disparate as well. Nevertheless, the marginal distributions of the inputs $P(X)$ are equivalent.

The other case of transfer learning is when $\mathcal{T}_S = \mathcal{T}_T$, but $\mathcal{D}_S \neq \mathcal{D}_T$, i.e., the tasks are the same, yet, the domains are different. Thus, the marginal probability distributions $P(X)$ and, possibly, the conditional probability distributions $P(Y \mid X)$ differ in the source and target domains. These phenomena are the

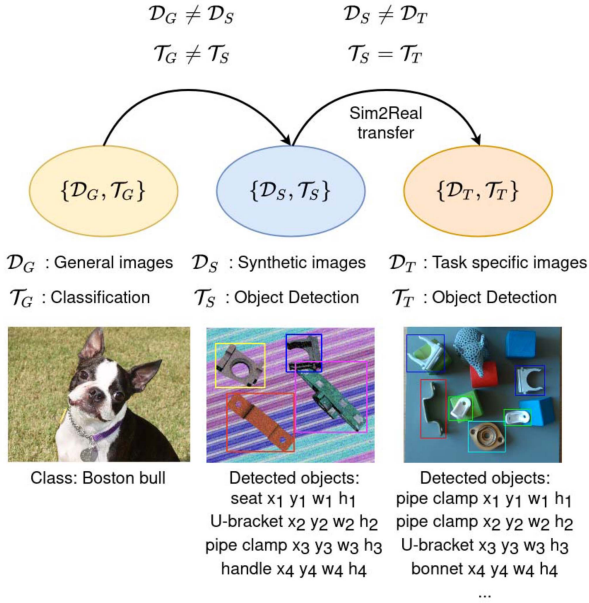

Fig. 1. Different phases of knowledge transfer. The picture of the Boston bull on the bottom left corner of the figure is from ImageNet [4].

*frequency feature bias*, and the *context feature bias*, respectively. In order to improve $f_T(\cdot)$, our aim is to extract the relevant (not domain-specific) knowledge from $\{\mathcal{D}_S, \mathcal{T}_S\}$. Thus, the model will perform well on the target domain ($\mathcal{D}_T$). If $\mathcal{D}_S$ and $\mathcal{D}_T$ are not related enough, negative transfer can occur, and knowledge transfer does not improve the performance of $f_T(\cdot)$, or even decreases it.

Sim2real object detection is a special case of transfer learning: instead of real images obtained from the target domain, the model is trained on synthetic data, thus $\mathcal{D}_S \neq \mathcal{D}_T$. On the other hand, it performs the same task, namely object detection (on the same classes of objects), therefore $\mathcal{T}_S = \mathcal{T}_T$. Nevertheless, the model trained on synthetic data, ceteris paribus, does not work on real images as the domains are disparate. This phenomenon is referred to as the *reality gap*, and the main goal of sim2real transfer is to bridge this gap. In our case, the sim2real transfer is the second phase of the knowledge transfer, shown in Fig. 1.

Domain adaptation (DA) is an approach to diminish the reality gap. It attempts to transform one domain into the other domain or transform both domains into a common domain. In the case of sim2real object detection, it usually consists in generating photo-realistic images for the training dataset. The more the generated images resemble the real ones, the more the difference between domains is reduced, and the performance with real images is improved. Typical data generation models for domain adaptation are based on variational autoencoders (VAE) [16] or generative adversarial networks (GAN) [17].

Domain randomization (DR), on the other hand, introduces variability by adding artificial noise to the synthetic training images. The idea is that noise makes the model robust to different domains, as it does not overfit on the domain-specific characteristics, but learns the domain-independent underlying data representation. Another possible interpretation is to regard the different domains as perturbed versions of one common domain. The general idea of introducing variance to simulation was first presented by Jacobi [18].

Other important concepts of transfer learning are the zero-shot and the one-shot transfers. In the context of object detection, zero-shot transfer means that not even one image is used from the target domain for training. In the case of one-shot transfer, only one or a few images are used from the target domain. A way to do one-shot transfer is to train the model on synthetic data and then fine-tune it on some examples from the target domain. In our one-shot transfer case, only one real image was used for training. Moreover, we did not separate the process into training and fine-tuning, as we mixed the copies of the real image and the synthetic images.

## III. RELATED WORK

In this section, the related work in sim2real knowledge transfer is presented in Section III-A. Furthermore, the different types of object detection models are presented in Section III-B.

### A. Domain Randomization and Domain Adaptation

Tobin et al. [14] trained a modified version of VGG-16 [19] deep neural network architecture for object localization. They generated nonrealistic synthetic RGB images randomizing the number and shape of the distractor objects, the position and texture of all objects, the texture of the background, the position, orientation and the field of view of the camera, the number of lights in the scene, the position, orientation, and specular characteristics of the lights, and the type and amount of random noise added to images. The random textures were either a random single color, a gradient between two colors, or a checker pattern of two random colors. The following nonindustrial objects were used: cones, cubes, cylinders, hexagonal prisms, pyramids, rectangular prisms, tetrahedrons, and triangular prisms. The images were rendered with the built-in renderer of the MuJoCo Physics Engine [20], and no real images were used for training the model. They achieved around 1.5 cm accuracy in the real-world environment. Tobin et al. [21] conducted further research, where they trained a deep neural network for grasp planning using only synthetic images and domain randomization, and achieved an 80% success rate in a real-world environment.

Borrego et al. [22] presented a plug-in for the Gazebo simulator [23]. Introducing variation reduced the reality gap between simulated and real-world data. In the case study, three types of objects were detected—box (cube), cylinder, and sphere. The simulated scenes contained a ground plate and a single light source. The objects were placed on a grid to prevent collusion, but they were rotated randomly. (In this regard, we found that introducing some disturbance to object placement significantly increases the performance, see in Section VIII-D.) Then, the camera and the light source were moved to random positions. Four different types of textures were used, namely, flat, gradient, checkerboard, and Perlin noise [24]. For training, the SSD [7], and separately, the faster R-CNN [6] networks were used. With the two networks, 70% and 88% $mAP_{50}$ were achieved, respectively, using 121 real images. Training the same networks with 9000 simulated images yielded 64% and 82% $\text{mAP}_{50}$, respectively. Interestingly, the hybrid approach (using real and synthetic images) accomplished 62% and 83% $\text{mAP}_{50}$, respectively. For all experiments, IOU 0.5 threshold (50 in $\text{mAP}_{50}$) was used, and the test results were validated on 121 real images (different from the 121 images used for the training). A follow-up ablation study [15] revealed that Perlin noise has a crucial influence on the performance of the model. Furthermore, data generation process was further accelerated to 9000 full-HD images in roughly 1.5 h (around 0.6 s per image).

Pashevich et al. [12] trained manipulation policies in a simulation environment with an object localization proxy task. Depth images for training were simulated in PyBullet [25] and gathered with a Kinect-1 camera from the real world. For finding the best data augmentation transformation and their order, Monte Carlo Tree Search (MCTS) [26] was used. The transformations were selected from the Python Image Library (PIL). The transformations were evaluated individually and as a sequence as well. From all transformations examined, the cutout transformation [27] performed best on real images (although in our experiments this was not the case for RGB images, see in Section VIII-E), and the best sequence of transformations was: cutout, erasing an object, white noise, edge noise, scale, salt noise, posterize, and sharpness, in this order. With the aforementioned sequence, $1.09 \pm 0.73$ cm position error was achieved in the real environment, for cubes of 4.7 cm edge length.

James et al. [28] trained an end-to-end robotic controller on synthetic data with domain randomization. The inputs of the deep-neural-network-based model were an image and the joint angles of the robot, while its output were motor velocities. The task was an abstract tidying manipulation, namely, putting a cube into a basket. Similarly to [22], Perlin noise was used as a perturbation. The model was examined in dynamically changing illumination settings, in the presence of distractors, including human presence, new cube size in test time, and with a moving basket. Experiments yielded at least a 75% success rate in all conditions, except for a spotlight and a smaller cube in test time. In these cases, the model had a 56% and a 41% success rate, respectively.

Devo et al. [29] used domain randomization to train a target-driven visual navigation model. The goal was to find a specific object in a maze. Maze wall heights, maze wall textures, maze floor textures, light color and intensity, and the light source angle were subject to randomization. For simulation, the Unreal Engine 4 [30] was used. An average of 72% success rate was achieved in simulation, and 46% in the real world. The experiments showed that direct sim2real transfer is possible for this kind of problem as well.

Chen et al. [31] created the domain adaptive faster R-CNN model for cross-domain object detection. Domain shift stemming from image-level and instance-level shifts were tackled with an approach based on $\mathcal{H}$-divergence theory and adversarial training. The study focuses on street images for self-driving cars where the domains are disparate due to different camera types and setups, different cities and diverse appearance of objects, or the particular weather conditions. Some experiments were also carried out with sim2real knowledge transfer, as the model was trained on SIM 10k [32] and evaluated on the Cityscapes dataset [33]. Their method improved the performance of the faster R-CNN model from 30.12 $\text{AP}_{50}$ to 38.97 $\text{AP}_{50}$ in the case of the car class.

Focusing on street scenarios, Sankaranarayanan et al. [34] proposed an unsupervised domain adaptation approach based on generative adversarial networks for semantic segmentation problems. For the synthetic source domain, the SYNTHIA [35] and the GTA V [36] datasets, and for the target domain, Cityscapes dataset [33] were used. The approach achieved 36.1 and 37.1 mIOU scores transferring knowledge from SYNTHIA and GTA V, respectively. Without domain adaptation, the method scored 26.8 and 29.6 mIOU.

Tremblay et al. [37] generated synthetic images with domain randomization techniques to perform object detection of cars in street scenarios. 100 K images were generated with maximum 14 cars each, selected randomly from 36 car models. The models were evaluated on the KITTI dataset [38]. Three DCNN architectures were trained (faster R-CNN [6], R-FCN [39], and SSD [7]), scoring 78.1, 71.5, and 46.3 $\text{AP}_{50}$, respectively, on the single-class object detection problem. Interestingly, better results were obtained than by training the same architectures on the virtual KITTI dataset [40] which has a high correlation to the KITTI dataset. The performance could be improved by fine-tuning the models on real images. With 6000 real images, the performance of the faster R-CNN reached 98.5 $\text{AP}_{50}$.

Hinterstoisser et al. [41] inserted 3-D models of objects in real images, using OpenGL with Phong shading [42] for rendering. Small perturbation were permitted in the ambient, diffuse, and specular parameters, and the light color. Gaussian noise and a blur with Gaussian kernel were added to better integrate the objects with the background. A faster R-CNN model was primarily used for training, with freezing the weights of the feature extractor. The latter significantly improved the performance of the model (although, Tremblay et al. [37] later reported the opposite effect in their case).

Zhang et al. [43] propose an adversarial discriminative sim2real approach to transfer visuo-motor policies. The method was demonstrated in a table-top object-reaching task. A blue cuboid object had to be reached with a velocity-controlled 7-DoF robot arm. The method could reduce the real data requirement by 50%, while 97.8% success rate and 1.8 cm control accuracy were achieved.

Clever et al. [44], [45] proposed a method to predict human position (resting on a bed) and contact pressure from depth data and gender information. The method achieved 3.837 $\text{MSE(kPa}^2)$ trained on 97 K synthetic images. In comparison, the same method reached 3.151 $\text{MSE(kPa}^2)$ trained on 11 K real images and 2.849 $\text{MSE(kPa}^2)$ trained on both real and synthetic images (108 K). For evaluation, the SLP dataset [46] was used.

Gomes et al. [47] proposed a simulated model for the GelSight tactile sensor. Having computed the height map of the elastomer, the internal illumination of the elastomer is calculated. The usefulness of the model was also demonstrated with a sim2real classification task. For the study, 12 texture maps resembling real objects were created and randomly perturbed on the captured synthetic data, improving the classification accuracy from 43.76% to 76.19%.

TABLE I
SUMMARY OF RELATED WORKS

| Work | Problem | Input | Domain | DR/DA | Base model | Simulator | Synthetic images | Real images | Results |
|------|---------|-------|--------|-------|------------|-----------|------------------|-------------|---------|
| Tobin [14] | Detection | RGB | Shapes | DR | VGG-16 | MuJoCo [20] | 5K–50K | 0 | 1.5 cm |
| Tobin [21] | Grasping | Depth | YCB [48] | DR | CNNs | MuJoCo [20] | 2K / obj | 0 | 80% success |
| Borrego [22] | Detection | RGB | Shapes | DR | Faster R-CNN | Gazebo [23] | 9K | 0 | 82% $mAP_{50}$ |
| | | | | | | | 9K | 121 | 83% $mAP_{50}$ |
| | | | | | SSD | | 9K | 0 | 64% $mAP_{50}$ |
| | | | | | | | 9K | 121 | 62% $mAP_{50}$ |
| Pashevich [12] | Detection | Depth | Cubes | DR | ResNet-18 | PyBullet [25] | 2K | 0 | 1.09±0.73 cm |
| | Cup placing | | Cups | | | | — | 0 | 15/20 |
| James [28] | Pick-and-place | RGB, joints | Cube | DR | CNNs | V-REP [49] | 100K–1M | 0 | ≥41% success |
| Devo [29] | Navigation | 2xRGB | — | DR | CNNs | UE4 [30] | 540K | 0 | 46% success |
| Chen [31] | Detection | RGB | Street | DA | Faster R-CNN | — | 10K | 3K (unlab.) | 38.97 $AP_{50}$ |
| Sankarana-rayanan [34] | Segmentation | RGB | Street | DA | GAN | — | 25K | 5K (unlab.) | 37.1 mIOU |
| Tremblay [37] | Detection | RGB | Street | DR | Faster R-CNN | UE4 [30] | 100K | 0 | 78.1 $AP_{50}$ |
| | | | | | | | 100K | 6K | 98.5 $AP_{50}$ |
| | | | | | R-FCN | | 100K | 0 | 71.5 $AP_{50}$ |
| | | | | | SSD | | 100K | 0 | 46.3 $AP_{50}$ |
| Hinter-stoisser [41] | Detection | RGB | Household | DR | CNNs | OpenGL | 20K | 0 | ca. 70% $mAP_{50}$ |
| Zhang [43] | Grasping | RGB | Cube | DR | VGG-16 | V-REP [49] | 3K | 93 + 186 (unlab.) | 97.8% success, 1.8 cm accuracy |
| Clever [44], [45] | Contact pressure | Depth | Humans | DR | CNNs | DART [50], FleX [51], Pyrender [52] | 97K | 0 | 3.837 $MSE(kPa^2)$ |
| | | | | | | | 97K | 11K | 2.849 $MSE(kPa^2)$ |
| Gomes [47] | Classification | RGB | Shapes | DR | ResNet-50 | Gazebo [23] | 1470 | 0 | 76.19% accuracy |
| **Our results** | **Detection** | **RGB** | **Industrial** | **DR** | **YOLOv4** | **PyBullet [25]** | **4K** | **0** | **86.32% $mAP_{50}$** |
| | | | | | | | **2K** | **1** | **97.38% $mAP_{50}$** |

The above works are diverse in terms of the problem itself, the input type, the domain of the application, and the amount of synthetic and real images used to train the model, making a complete comparison a challenge. Nevertheless, a general overview organized by selected characteristics is presented for reference in Table I. In general, certain limitations of the above works relate closely to our work (solving object detection).

1) The classification part of the problem is less challenging as the works use easy shapes such as cubes, spheres, and cones, or even have one class only.
2) The works rely on considerably more synthetic and real images for training.

Even though the cited works use transfer learning (domain adaptation or domain randomization) to reduce the reality gap, they do not solve the same machine learning problem, and may use different models as well. All of this needs to be taken into consideration if an in-depth comparison is desired.

### B. Object Detection

A comprehensive overview of object detection models and the history of the field is not in the scope of this article, therefore, we limit this section to a selection of sources relevant to our work.

For further detail, we refer the reader to standard surveys such as the work of Zou et al. [53].

DCNN architectures can be categorized into two groups: two-stage detectors and one-stage detectors. Two-stage detectors have a proposal detection stage where a set of bounding box candidates is generated, and a verification stage where these bounding boxes are separately evaluated whether they contain an object of a specific class. Examples of these networks are R-CNN [54], SPPNet [55], fast R-CNN [56], faster R-CNN [6].

In the case of one-stage detectors, on the other hand, a single neural network is applied to the full image that predicts the bounding boxes straight away. The slow detection time, which is the biggest disadvantage of the two-stage detectors, can be overcome with the one-stage approach. Detection time is crucial for many applications, especially but not exclusively in the field of robotics or self-driving cars. Redmon et al. proposed the first one-stage detector YOLO in 2015 [57], being the first real-time object detector. Subsequent updates introduced its second [58] and third versions [59]. Single shot multibox detector (SSD) [7] and RetinaNet [60] are two other popular one-stage detectors.

Bochkovskiy et al. [8] created the fourth version of YOLO aiming to improve the accuracy of the model while still keeping an optimal accuracy–speed tradeoff. With the CSPDarknet-53 [8] backbone, 65.7% $\mathrm{mAP}_{50}$ could be achieved for the MS COCO dataset [5] and around 65 FPS speed on a Tesla V100. In comparison, on the same dataset, SSD with VGG-16 [19] backbone performed 48.5% $\mathrm{mAP}_{50}$ and RetinaNet with ResNet-101 [61] backbone achieved 57.5% $\mathrm{mAP}_{50}$.

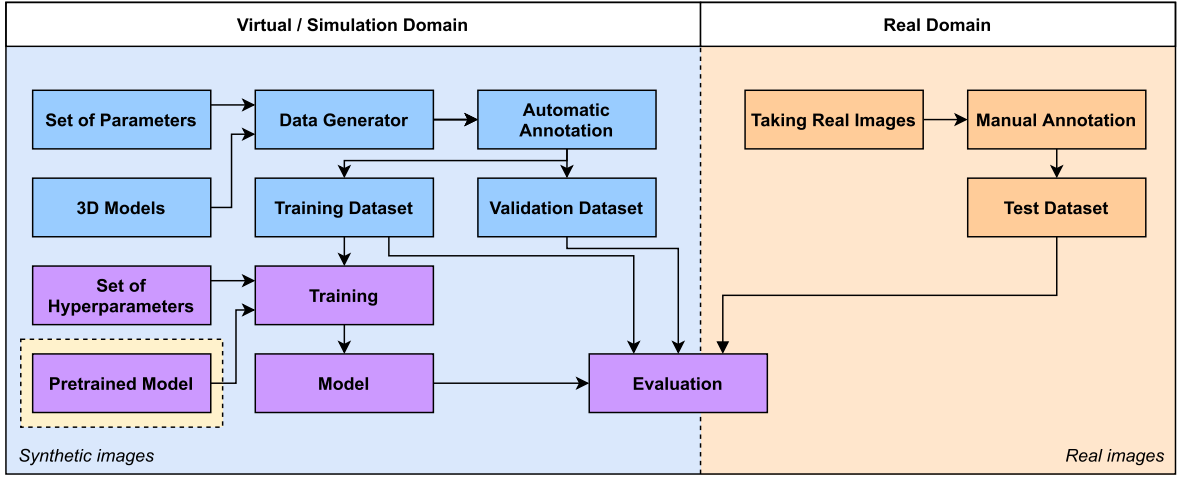

Fig. 2. Flowchart diagram of our data generation, training, and evaluation process. The blue and orange boxes depict the data generating and data gathering steps. The purple boxes represent the steps of training and evaluation.

## IV. METHOD

This section presents our method in detail—namely, the proposed sim2real knowledge transfer in Section IV-A, the data generation module in Section IV-B, and the training module in Section IV-C. The implementation is freely available at[3].

### A. Sim2Real Knowledge Transfer

The flowchart diagram of our data generation, training, and evaluation process is depicted in Fig. 2. It can be broken down into functionally separable tasks. The data generation process creates randomized and postprocessed synthetic images of given objects. It also automatically generates the annotations for the images. Thus, the output of the data generation process is a set of images paired with their labels grouped into a training and a validation dataset.

In order to train the model (with the set of hyperparameters), only the images from the training dataset are used. As the initial layers of the neural network perform low-level image processing tasks such as detecting contours, lines, or edges, we utilized a pretrained image classifier model as a feature extractor of our object detector. This is the first phase of our knowledge transfer, depicted on the left side of Fig. 1. The knowledge transfer goes from $\{\mathcal{D}_G, \mathcal{T}_G\}$, where $\mathcal{D}_G$ is the domain of the dataset of general public images and $\mathcal{T}_G$ is classification, to $\{\mathcal{D}_S, \mathcal{T}_S\}$, where $\mathcal{D}_S$ is the domain of synthetic images (source domain of the sim2real knowledge transfer), and $\mathcal{T}_S$ is object detection. Here, $\mathcal{D}_G \neq \mathcal{D}_S$, and $\mathcal{T}_G \neq \mathcal{T}_S$. The second phase of knowledge transfer is the sim2real transfer which goes from $\{\mathcal{D}_S, \mathcal{T}_S\}$ to $\{\mathcal{D}_T, \mathcal{T}_T\}$, where $\mathcal{D}_T$ is the domain of our industrial environment (target domain), and $\mathcal{T}_T$ is object detection. Here, $\mathcal{D}_S \neq \mathcal{D}_T$ but $\mathcal{T}_S = \mathcal{T}_T$. Although the pretrained network does possess some learned knowledge from the domain of a given general public dataset ($\mathcal{D}_G$), it does not have direct knowledge of the target objects. Consequently, $\mathcal{D}_G \neq \mathcal{D}_S \neq \mathcal{D}_T$. Even though $\mathcal{D}_G \neq \mathcal{D}_T$, the characteristics of the domains are similar. Thus, in Figs. 1 and 2, the domains of general public images and the task-specific real images are marked with different shades of orange.

### B. Data Generation

The data generation process is responsible for the creation of synthetic images paired with accurate automatic ground-truth annotations. In several stages of this process, artificial random perturbations are applied as domain randomization techniques.

1) *Framework:* For data generation, the PyBullet [25] physics simulator was utilized since it has an easy-to-use, intuitive API, including an image renderer tool, and an integrated physics simulator where the gravitational force can be simulated easily.

The duration of dataset generation is a relevant aspect of the method, as in the industry, on many occasions, it is not feasible to wait long hours or even days to start the training, which can be a time-consuming process itself. One of the biggest advantages of domain randomization over domain adaptation is that it is generally faster as images do not need to be photo-realistic. In our case, for data generation, we could achieve less than 0.5 s per image on a GeForce RTX 2080 Ti GPU. With 4000 images, this amounts to around 33 min. If one image is rendered in 1 min instead of the 33 min (which is plausible in the case of photo-realistic images), the aforementioned 4000 images would take more than 66 h. Having generated the dataset, the training lasts around 12 h, thus, a complete generation and training process can be executed automatically in around 13 h.

2) *Object Generation:* The framework is capable of placing any type of object into the simulation if its 3-D description file is given. In the case of industrial applications, which is the aim of this research, these 3-D models are easily accessible.

TABLE II
MOST RELEVANT INPUT PARAMETERS OF THE DATA GENERATOR MODULE IN TERMS OF OBJECT GENERATION

| Notation | Type | Description |
|---|---|---|
| $n$ | Scalar | The grid size ($n \times n$). |
| $d$ | Scalar | The grid spacing. The distance between the adjacent grid points. |
| $z_{pos}$ | scalar | The position of an object in $z$ direction. |
| $\epsilon_{pos}$ | array | Contains the proportionate (to the grid spacing) perturbation limits in the directions of $x$, $y$, and $z$ around the center point. |
| $\epsilon_{rot}$ | List of arrays | $\epsilon_{rot}$ describes the lower and upper bounds for the possible rotation angles of the directions in $x$, $y$, and $z$. |
| $P_{objects}$ | Array | Contains the selection probabilities of the given objects (including the distractor object and the void object). |
| $P_{texture}$ | Scalar | The probability that a specific object or the ground plane is having a random texture. Otherwise, it gets a random color. |

The most relevant input parameters of the object generation process are summarized in Table II and the process works as follows.

- A horizontal plane is placed at the vertical $z = 0$ position.
- According to the grid size ($n$) and the grid spacing ($d$) parameters, the $x$ and $y$ coordinates of the grid points are set.
- The vertical $z > 0$ coordinates of the grid points are set by $z_{pos}$.
- The objects are not placed exactly at the grid points. The $x$, $y$, and $z$ coordinates of the objects are obtained from a uniform distribution in which the center point is a given grid point and the limits of the distribution are set by $\epsilon_{pos}$.
- Once object selection has been performed, the appropriate 3-D model of the object is loaded into the specific coordinates. Predefined weights describe the probability of selecting a specific object. Furthermore, a distracting cuboid object (which is not in any of the classes) or a void object (leaving that grid point empty) can be selected. The sizes of the distracting objects are also individually randomized. The aforementioned probabilities are set by $P_{objects}$.
- The objects are also randomly rotated around the $x$, $y$, and $z$ axes, described by a uniform distribution whose limits are set by $\epsilon_{rot}$.
- The objects and the ground plane are given some random textures drawn from three public datasets [62], [63], [64], with the probability of $P_{texture}$. Some examples of the textures are shown in Fig. 3. Random RGB colors are assigned to the objects (or to the ground plane) which do not receive any texture.
- Before rendering the image, the objects are dropped down from $z > 0$ to the ground plane. Thus, the objects are captured in a natural stable position. The simulation of the free fall takes around 0.05–0.1 s per scenario (with every step included, the generation of an image with its label is around 0.45–0.5 s).

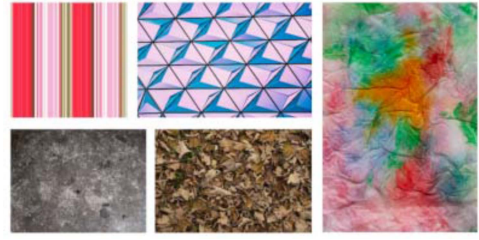

Fig. 3. Some examples of the textures used [62], [63], [64].

TABLE III
MOST RELEVANT INPUT PARAMETERS OF THE DATA GENERATOR MODULE IN TERMS OF IMAGE RENDERING

| Notation | Type | Description |
|---|---|---|
| $T_{pos}$ | List of arrays | The camera points to a certain point in the 3-D space. The arrays of $T_{pos}$ describe the lower and upper bound of the camera target point in the directions of $x$, $y$, and $z$. |
| $C_{pos}$ | List of arrays | The pose of the camera is defined by three parameters: the rotation angle around the global z-axis, the pitch angle from the global $x$–$y$ plane, and the distance between the camera and the target point. The arrays of $C_{pos}$ describe the lower and upper bound of these parameters. |
| $R_{width}$ | Array | The $R_{width}$ parameter describes the lower and upper bound of image width in pixels. |
| $R_{height}$ | Array | The $R_{height}$ parameter describes the lower and upper bound of image height in pixels. |
| $FOV$ | Scalar | The field-of-view of the camera. |
| $I_{type}$ | Scalar | Three types of images can be rendered: 3-channel RGB images, depth images, and 4-channel RGB-D images. |

*3) Image Rendering:* For rendering an image, the camera pose, its inner parameters, and additional parameters must be set. The most relevant parameters of the image rendering are presented in Table III. The algorithm works as follows.

- The camera is placed in a randomized position pointing to a random point around the center of the grid defined by $C_{pos}$ and $T_{pos}$. The randomization is constrained to ensure that the center points of all objects appear on the rendered image.
- The camera field-of-view (FOV) and its additional intrinsic parameters are set. Image width and height are obtained from a uniform distribution defined by $R_{width}$ and $R_{height}$.
- The RGB, D (depth), or RGB-D images are taken, defined by $I_{type}$. RGB-D images are created by concatenating the RGB and the D images. For one layout (object generation), only one image is taken.

*4) Label Generation:* Having generated the objects and rendered the image, the ground-truth bounding box (BB) parameters must be computed. For different CNNs, the format is different, but it is generally true that the bounding box parameters describe all the objects in an image as follows.

- The center point ($x$ and $y$) of the object.
- The width and the height of the bounding box.
- The class of the object.

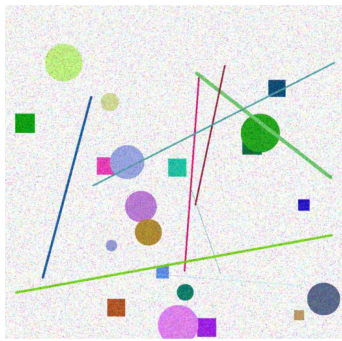

Fig. 4.    Postprocess transformation on a blank image.

| Name | Description |
|---|---|
| Multicolor pepper-and-salt | With a given $p$ probability, every channel of every pixel is set either to zero or to the maximum value. |
| Gaussian blur | With a given $p$ probability, the whole image is blurred with a randomized filter size. |
| Cutout | A given number of rectangle-, circle-, or line-shaped regions with randomized dimensions and at randomized positions are colored to a random RGB value. |

| Name | Description |
|---|---|
| Angle | Randomly rotates images. |
| Saturation | Randomly changes the saturation. |
| Exposure | Randomly changes the brightness. |
| Hue | Randomly changes the hue color channel. |
| Mosaic | Use four images in one Mosaic data augmentation technique [8]. |
| Jitter | Randomly changes the size of the images and their aspect ratio. |
| Random | Randomly resizes network size after every ten batches. |

The aforementioned ground-truth generation is an automatic process involving a coordinate transformation from the simulation 3-D world coordinate system to the image 2-D coordinate system.

The 4×4 transformation matrix is the matrix product of the view matrix and projection matrix of the camera, respectively. In order to transform a point from the world coordinate system to the image space, it must be multiplied with this transformation matrix and then scaled back by its fourth coordinate to get the true projection. For a detailed explanation of the projection, we refer the reader to [65].

In the framework, we implemented two ways of computing the bounding boxes. The two approaches differ in the number of points that are transformed into the image space. One approach transfers only the eight points (per object) of the world 3-D axis-aligned bounding boxes (AABB), which is available in the PyBullet simulator, whereas the other transforms all the points of the objects to the image space. Henceforward, we refer to the former approach as the eight-point method and the latter as the all-point approach. Having obtained the transformed points, the second part of the algorithm is the same in both cases: the minimum and maximum values in $x$ and $y$ directions are selected to define the limits of the BBs. The center point can be computed as the arithmetic means of the minimum and maximum values. As a result, the latter method gives tighter, more accurate bounding boxes at the cost of extra calculations.

*5) Postprocessing:* The technique of domain randomization was performed in multiple steps of the previously defined synthetic image generation process. In the postprocessing phase, as a domain randomization technique, additional artificial noise is introduced to alter the images. The images are perturbed with a randomized multicolor pepper-and-salt noise and a Gaussian blur. Furthermore, as Pashevich et al. [12] found the rectangle cutouts useful for depth images, experiments were made with rectangle cutouts, and additional circle, as well as line cutouts in our RGB images. The noise types are shown in Fig. 4, described in Table IV, and were applied in the following order:

1) rectangle cutout;
2) circle cutout;
3) line cutout;
4) multicolor pepper-and-salt;
5) Gaussian blur.

The goal of postprocessing is to force the model not to learn the synthetic domain-specific characteristics, but to try to learn the domain-independent underlying data representation. The ablation study on our experiments, described in Section VIII, shows that having the postprocessing module undoubtedly improves the performance of the models with the test dataset. Nevertheless, it also reveals that the added cutout noises did not improve the performance compared to the default Gaussian blur and multicolor pepper-and-salt noise.

*C. Training*

Even though the method would work with any given CNNs, we have chosen the YOLOv4 [8] architecture for this research for the following reasons.

1) It has the best speed and accuracy tradeoff which makes it a good fit for robotic applications. It also has a tiny version, allowing it to run in real-time even on a microcomputer such as a Raspberry Pi or NVIDIA Jetson Nano.
2) Its training framework contains additional advanced data augmentation tools. For more information, we refer the reader to [8]. These tools help to introduce further perturbation to the system.

For the training, a model pretrained on ImageNet [4] is used. The most relevant hyperparameters for the advanced data augmentation tools are shown in Table V, keeping the original names of the parameters.

In Section VIII-F, we present the results of our method only changing the object detection model from YOLOv4 to faster R-CNN.

## V. Evaluation

In this section, we outline the metrics used to evaluate our models. First, we define how we measured the reality gap, then we introduce our altered version of the confusion matrix, and last, we outline some further details of our evaluation process.

To evaluate the solution of the classical machine learning problem, (training and evaluation on the same domain), real-world images would not be needed. In this case, the performance is assessed on the generated validation dataset that is not used for training but comes from the same distribution $P_{train}(X) = P_{valid}(X)$. The solution can be assessed by the value of the mAP score of the model on the valid dataset, and the capability of generalization (within the specific domain) can be measured by comparing the performance of the model on the training and the validation datasets, as in the following:

$$G_{ML} = mAP_{train} - mAP_{valid}. \tag{1}$$

To evaluate the performance of the knowledge transfer, a manually annotated test set of real images is needed. In this case, $P_{valid}(X) \neq P_{test}(X)$. We expect that the given model performs notably worse on the test set than on the validation and training sets. To measure the magnitude of the reality gap, we can define it as the difference of the performance of the model on the validation and test sets, as shown in the following:

$$G_{reality} = mAP_{valid} - mAP_{test}. \tag{2}$$

Furthermore, we adapt the confusion matrix measure from the field of image classification to object detection and use it as an additional performance measure. The adaptation works as described as follows.

1) Adding an extra row and an extra column to the classical confusion matrix. Thus, there are $c + 1$ rows and columns, where $c$ is the number of classes. The additional column represents the objects that are not predicted to any of the classes but actually belong to one class. On the other hand, the additional row of the matrix represents the cases when the model predicted an object of a class in a position where there should not be any object.

2) The values of the diagonal are the correct predictions. For simplicity, the last element of the diagonal is zero. This element should contain the number of objects that are not in the images and the model rightfully did not find them, which does not have any meaning.

3) As more than one prediction can belong to one ground-truth object, a given ground-truth object appears in the matrix as many times as many predictions are paired with it. Therefore, contrary to the traditional confusion matrix, the sum of all elements in the matrix will not necessarily be equal to the sum of all ground-truth objects or predictions.

The adapted confusion matrix described is proven to be useful for detecting and quantifying misclassifications which turned out to be the primary cause of performance loss in the case of similar classes. Examples are shown in Section VII in Figs. 14 and 20.

As presented in detail in Section VII, several training runs were carried out to test our method. For every dataset generated,

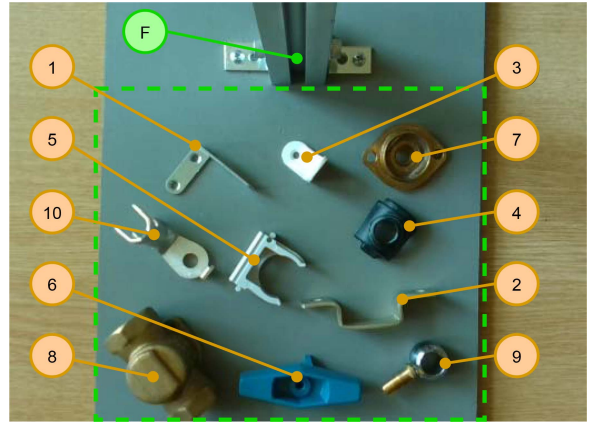

Fig. 5. Selected industrial parts for the dataset. Their names in order of their identifier numbers are the following: 1. L-bracket, 2. U-bracket, 3. angle bracket, 4. seat, 5. pipe clamp, 6. handle, 7. bonnet, 8. body, 9. ball, 10. cable shoe. The letter "F" designates the camera holder frame. The green dashed lines show the borders of the cropped images.

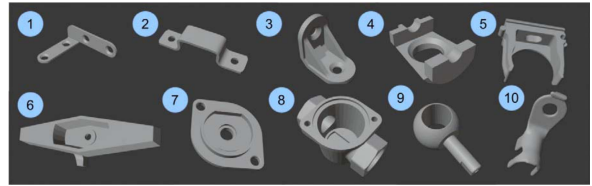

Fig. 6. Three-dimensional models of the selected industrial parts of the dataset. Their names in order of their identifier numbers are the following: 1. L-bracket, 2. U-bracket, 3. angle bracket, 4. seat, 5. pipe clamp, 6. handle, 7. bonnet, 8. body, 9. ball, 10. cable shoe. Their scaling factors are different for better visualization.

three independent training sessions were conducted, resulting in three different models (sets of weights) in order to measure the deviance of the training process. The average performance of the models refers to the arithmetic mean of the results of these three models. We also use the $F_1$ score measure, which is defined as the harmonic mean of the precision and the recall values.

## VI. Dataset

This section presents the dataset created for the validation elaborated in detail in Section VII. Ten industrial parts were selected for the dataset. Object diversity as well as object similarity were the two major points of consideration. The former helps us to evaluate the detection performance of the model for various types of objects, whereas the latter is important in assessing the classification performance of the model. In general, it is easier to misclassify objects with similar features. Thus, this problem can be considered to be more challenging than the detection of less complex and fairly different shapes such as cubes and spheres. The selected objects are depicted on Fig. 5, and their virtual counterparts on Fig. 6. These images are samples of $X \in \mathcal{X}$ obtained from two different $P(X)$ probability distributions. The virtual images are from the probability distribution $P_S$ of $\mathcal{D}_S = \{\mathcal{X}, P_S(X)\}$, while the real images are from the probability distribution $P_T$ of $\mathcal{D}_T = \{\mathcal{X}, P_T(X)\}$.

As it can be seen, on one hand, objects of different sizes, shapes, colors, and materials were selected to increase diversity.

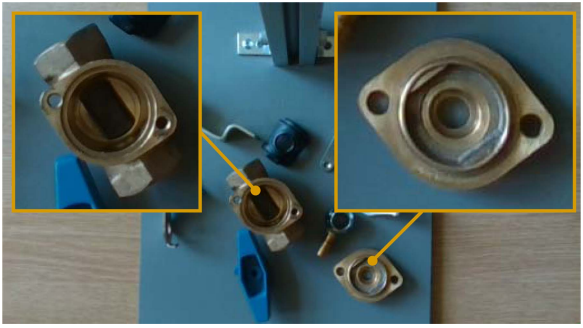

Fig. 7. Similarity of the body and the bonnet objects.

On the other hand, some objects share similar characteristics, such as circular holes. Furthermore, two parts, the bonnet (#7) and the body (#8) were chosen because of their high level of similarity, as shown in Fig. 7.

Constructing the dataset, 190 real images of 920 object instances were taken with different layouts and illumination settings. The images were captured with an Intel RealSense D435 camera. For easy and fast image capturing, a frame was designed that holds the camera 310 mm from the ground. A $300 \times 210 \times 10$ mm light blue wooden base supports the frame—this is also where the parts were placed. The images show not only this base area but the background (tabletop) as well—this is done on purpose. In order to have a slightly different dataset as a reference, we also transformed the aforementioned dataset by cropping the images to fit in the wooden base. The cropped area is signed with dashed green lines in Fig. 5. Some examples of cropped images are shown in Figs. 15 and 21.

The annotations for the test dataset were labeled manually and saved in the YOLO annotation format. As it contains all the necessary bounding boxes and class information, other annotation formats can be generated from them. We emphasize that these images of real objects were not used at any point for training the models, except in the case of one-shot knowledge transfer. In this case, only one real image was used. The experiment of one-shot transfer is presented in Section VII-B.

For all images, the matching depth images are recorded as well. The depth images are transformed in a way that each pixel point of the RGB image can be associated with the same pixel point of the depth image (the transformation is necessary as the fields-of-view of the cameras for RGB and for depth images are different). Thus, all the annotations for the RGB images are the same for the depth images. Even though the depth images were not used in the current research, this additional data can be valuable for later use or for other researchers.

The test dataset is summarized in Table VI. In Group A, every image contains only one object (except one image without any objects). In Group B, every image contains multiple objects, but no class is represented more than once. In Group C, spotlight illumination is applied from one side to test the robustness of the models to illumination settings. In Group D, cluttered scenes are recorded. In Group E, distractor objects (cubes, cylinders, triangular prisms, and a 3-D-printed elephant) are placed in the scene. Finally, in Group F, in every picture, only one class is presented, however, unlike in group A, there are several instances of this class in every image.

TABLE VI
SUMMARY OF THE TEST DATASET

| ID | #Img | #Obj | Object density | Illumination | Distractors |
|---|---|---|---|---|---|
| A | 53 | 52 | 0.98 | Normal | No |
| B | 19 | 182 | 9.58 | Normal | No |
| C | 20 | 144 | 7.20 | Spotlight | No |
| D | 21 | 264 | 12.57 | Normal | No |
| E | 22 | 135 | 6.14 | Normal | Yes |
| F | 55 | 143 | 2.60 | Normal | No |
| SUM | 190 | 920 | 4.84 | | |

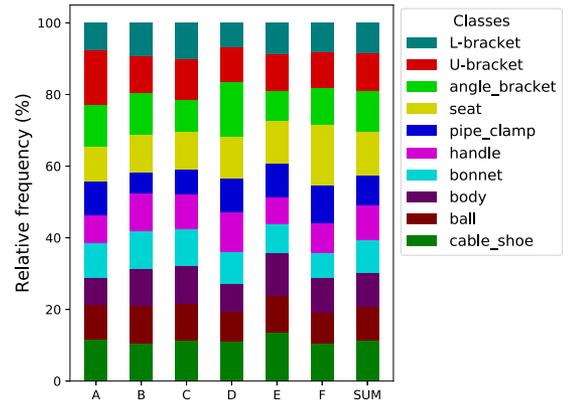

Fig. 8. Class distributions of the test dataset.

The group-wise class distributions are depicted in Fig. 8. As it can be seen, the classes are relatively equally distributed in the groups. Even though in the mAP metric, the mean of the classes is calculated, thus it is less influenced by class imbalance, it is advantageous to create an equally distributed test dataset. Obviously, for training, which can be sensitive to class imbalance, the synthetic data is generated with a random selection of objects, thus eliminating any notable class imbalance. The dataset can be downloaded from the project repository:[4]

## VII. RESULTS

In this section, we show the strengths of our sim2real object detection method, as described in Section IV, by applying it for the problem presented in Section VI.

### A. Zero-Shot Transfer (ZST)

The best-performing zero-shot transfer model ($\texttt{ZST\_BEST}_1$)[5] achieved 86.32% $\mathrm{mAP}_{50}$ on the cropped test dataset. For data generation, a $2 \times 2$ grid with fixed $z$ positions and a placement disturbance of $\pm 10\%$ of the grid spacing was set in the horizontal directions. The simulation of

---

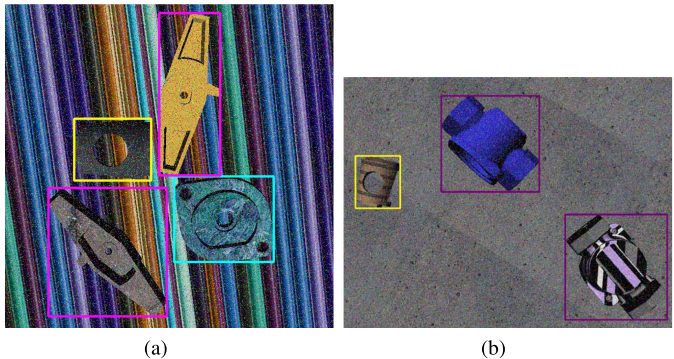

(a)   (b)

Fig. 9. Two examples of synthetic images with the automatically generated annotations. The bounding boxes are shown here for illustration purpose only. (a) Example 1. (b) Example 2.

gravity was enabled and the objects (including distractors and empty places) were selected with equal chance. The objects had random textures with a probability of 0.8 and a random color with a probability of 0.2. The camera target position was set to the center of the grid with a fixed 45° field-of-view. The pitch of the camera was randomized between −0.17 and 0.17 radians. The width and height of the images are chosen randomly, independently of each other. Their values lie between 640 and 1300 pixels. For postprocessing, multicolor pepper-and-salt noise and Gaussian blur were used with the probability of 1.0 and 0.5, respectively.[6]

With these parameters, 4000 images were generated for the training dataset, and 200 for the validation dataset. The evaluation of the model's performance on the training set was measured only on the first 200 images of the training set. Two examples of the synthetic images are shown in Fig. 9.

The precision-recall curves of the three `ZST_BEST` models (from the three training sessions) are shown in Fig. 10. As both the training and validation mAPs are close to 100%, it can be stated that the solution of the classical machine learning problem is satisfactory. Furthermore, observing the sim2real transfer, it can be seen that the models not only have a relatively good performance on the test data, but also have little variance. Moreover, the models perform relatively similarly on the original and on the cropped images which shows the robustness of the method. The $F_1$ score is depicted in Fig. 11. It shows that while the performance of the model on the cropped images is not affected by the threshold, the models work better with higher threshold values on the original images.

The performance of the models on the different groups of the test dataset (described in Table VI) is presented in Table VII. The performance is relatively stable across the different scene types. Group D, containing the most crowded images, performs just slightly worse than the others. In the case of the original images, group A has relatively low performance. This is due to the fact that the model occasionally falsely identifies the brackets of the camera holder frame (at the two sides of the

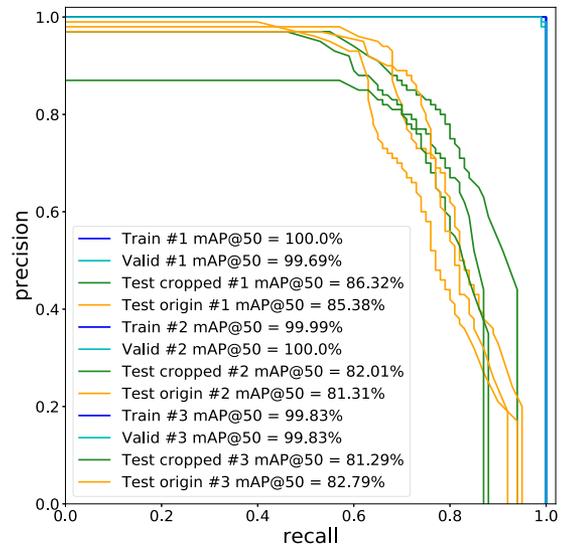

Fig. 10. Precision-recall curves of the `ZST_BEST` models. The train and valid scores are overlapping and relatively close to the perfect 1.0 values.

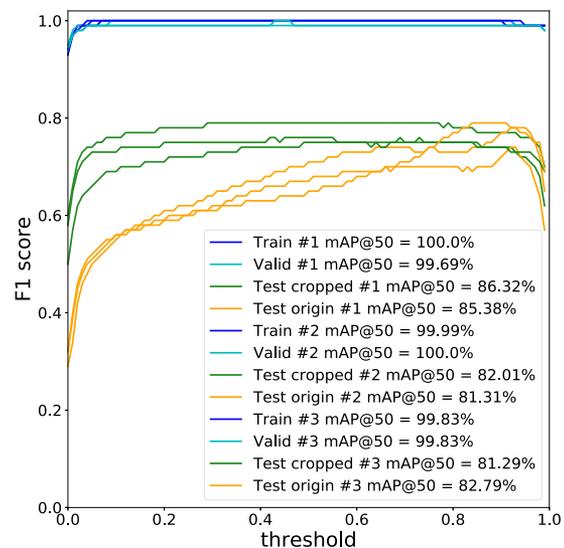

Fig. 11. $F_1$ scores of the `ZST_BEST` models.

TABLE VII
mAP$_{50}$ SCORES OF THE `ZST_BEST` MODELS IN THE DIFFERENT TEST GROUPS

| ID | Train #1 Orig. | Crop. | Train #2 Orig. | Crop. | Train #3 Orig. | Crop. | AVG. Orig. | Crop. |
|----|-------|-------|-------|-------|-------|-------|-------|-------|
| A | 82.43 | 95.71 | 75.85 | 84.86 | 83.97 | 87.00 | 80.75 | 89.19 |
| B | 89.27 | 86.48 | 85.70 | 80.26 | 86.00 | 81.18 | 86.99 | 82.64 |
| C | 85.02 | 82.41 | 86.69 | 82.27 | 83.97 | 79.11 | 85.23 | 81.26 |
| D | 84.50 | 83.94 | 82.54 | 80.07 | 79.89 | 76.58 | 82.31 | 80.20 |
| E | 87.22 | 85.05 | 82.99 | 84.91 | 82.02 | 83.52 | 84.08 | 84.49 |
| F | 86.03 | 95.11 | 79.63 | 87.50 | 88.41 | 91.68 | 84.69 | 91.43 |

[6] The cutouts did not improve the performance, as shown in Section VIII-E, thus, they were not used here.

TABLE VIII
mAP$_{50}$ Scores of the ZST_BEST Models for the Different Classes

| ID | Train #1 | | Train #2 | | Train #3 | | AVG. | |
|---|---|---|---|---|---|---|---|---|
| | Orig. | Crop. | Orig. | Crop. | Orig. | Crop. | Orig. | Crop. |
| 1 | 41.45 | 63.15 | 45.26 | 71.85 | 49.28 | 66.54 | 45.33 | 67.18 |
| 2 | 76.34 | 91.66 | 77.51 | 91.58 | 75.32 | 93.56 | 76.39 | 92.27 |
| 3 | 94.14 | 93.14 | 91.56 | 93.17 | 93.39 | 93.11 | 93.03 | 93.14 |
| 4 | 84.96 | 77.71 | 71.88 | 60.16 | 92.72 | 70.74 | 83.19 | 69.54 |
| 5 | 99.82 | 95.72 | 99.37 | 99.49 | 99.76 | 99.97 | 99.65 | 98.39 |
| 6 | 95.77 | 98.21 | 94.82 | 97.53 | 98.90 | 98.14 | 96.50 | 97.96 |
| 7 | 71.77 | 54.26 | 52.39 | 22.01 | 30.57 | 15.90 | 51.58 | 30.72 |
| 8 | 96.51 | 92.13 | 93.77 | 92.23 | 93.26 | 79.46 | 94.51 | 87.94 |
| 9 | 97.13 | 99.03 | 94.27 | 96.89 | 98.14 | 99.27 | 96.51 | 98.40 |
| 10 | 95.87 | 98.19 | 92.32 | 95.22 | 96.57 | 96.24 | 94.92 | 96.55 |

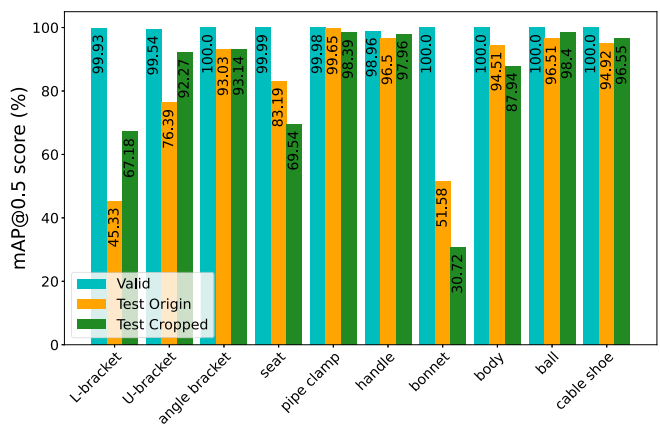

Fig. 12. Average mAP$_{50}$ scores of the ZST_BEST models in the different classes.

camera holder frame which is displayed in Fig. 5, marked with letter "F") as L-brackets—this is not surprising as these parts look similar. As group A has the lowest number of objects, this phenomenon has the most impact on results in this case. The cropped images, as shown in Fig. 5, do not contain this part of the image.

Furthermore, the performance of the models for the different classes is worth investigating. The data are presented in Table VIII and the average results are shown in Fig. 12. Looking at the dataset of cropped images (green), it can be seen that 6 out of 10 classes perform above 92%, one class is relatively close to them with 87.94% AP$_{50}$, two classes have worse results with 69.54% and 67.18% AP$_{50}$, and one class—the bonnet—has significantly worse performance with 30.72% AP$_{50}$. Otherwise, the performance on the validation dataset is close to 100% for all classes. The findings indicate that the performance loss is not caused by the unsuccessful solution of the classical machine learning problem, but by the existence of the reality gap. As most of the classes have relatively good APs, the bad classes are outweighed by them in the mAP calculation.

In order to investigate the aforementioned problem, the class-wise precision-recall graph of ZST_BEST$_1$ is shown in Fig. 13.

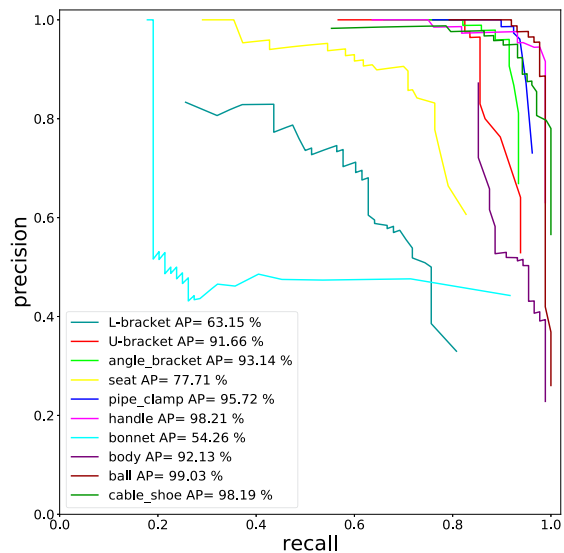

Fig. 13. Class precision-recall curves of the ZST_BEST$_1$ model on the cropped images.

Fig. 14. Confusion matrix of the ZST_BEST$_1$ model on the cropped images with the threshold set to 0.8.

As it can be seen, the bonnet, the L-bracket, and the seat are the worst classes consistently with Fig. 12.

The proposed confusion matrix depicted in Fig. 14 (described in detail in Section V) is essential in finding the root causes of the weaknesses of the models. In most cases, the objects are detected and classified to the correct class (diagonal). However, several instances of L-bracket and seat are not detected (36 and 49 examples) and many bonnets are classified as body objects (62 instances). The problem with these two objects is not surprising considering the high level of similarity of the two objects, as shown in Fig. 7. In general, this representation of the results not only confirms the aforementioned assumptions of class performances but also shows the underlying reason behind the lack of performance in their cases.

Finally, having presented the quantitative evaluation, two examples are given for qualitative evaluation as well. Fig. 15 shows an accurate and an inaccurate example, both with the threshold set to 0.8. While the model could accurately find and classify all the parts even in the presence of distractor objects in Fig. 15(a), it fails to detect two instances of the seat class and assigns the bonnet object to the body class in Fig. 15(b).

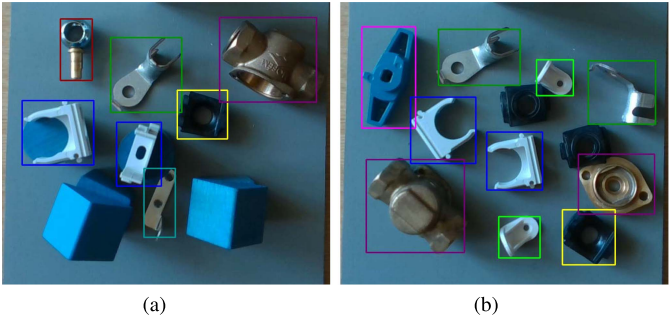

Fig. 15. Accurate (a) and an inaccurate (b) prediction of the `ZST_BEST`$_1$ model with the threshold set to 0.8. The color-coding follows Fig. 8. (a) Accurate example. (b) Inaccurate example.

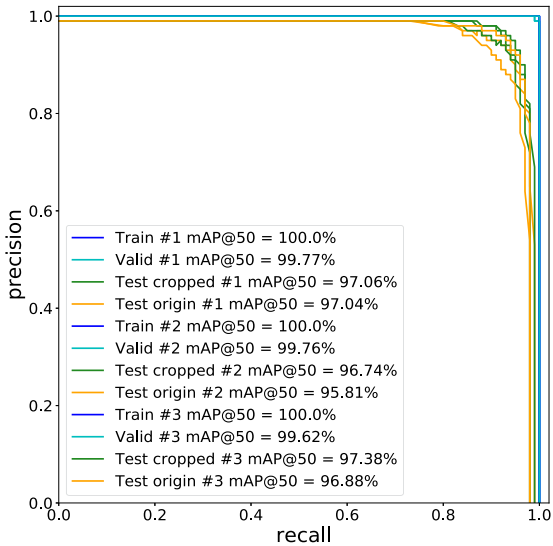

Fig. 16. Precision-recall curves of the `OST_BEST` models. The training and validation scores are overlapping and relatively close to the perfect 1.0 values.

TABLE IX
TRAINING DATASETS

|  | Synthetic images | Real images |
|---|---|---|
| ZST | 4000 | 0 |
| OST | 2000 | 1 (copied x2000) |

### B. One-Shot Transfer (OST)

Even though the best zero-shot transfer achieved 86.32% $mAP_{50}$, it had some difficulties with four classes. With one-shot transfer, we could overcome these difficulties. The hyperparameters of data generation remained the same as it was described in the previous zero-shot transfer example. The difference between the two approaches lies in the data used to train the model, which is shown in Table IX.

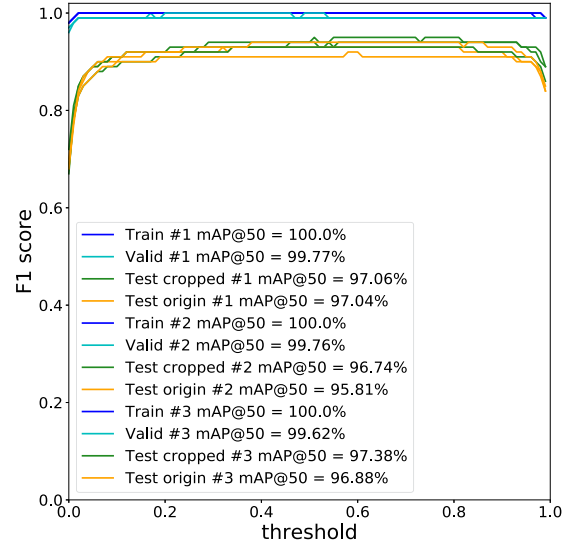

Fig. 17. $F_1$ scores of the `OST_BEST` models.

TABLE X
$mAP_{50}$ SCORES OF THE `OST_BEST` MODELS IN THE DIFFERENT TEST GROUPS

|  | Train #1 | | Train #2 | | Train #3 | | AVG. | |
|---|---|---|---|---|---|---|---|---|
| ID | Orig. | Crop. | Orig. | Crop. | Orig. | Crop. | Orig. | Crop. |
| A | 97.29 | 97.52 | 97.76 | 97.29 | 98.00 | 97.76 | 97.68 | 97.52 |
| B | 98.12 | 96.45 | 97.29 | 97.02 | 97.92 | 97.49 | 97.78 | 96.99 |
| C | 96.39 | 96.19 | 94.18 | 96.40 | 96.21 | 97.18 | 95.59 | 96.59 |
| D | 96.23 | 95.94 | 94.97 | 94.42 | 95.60 | 95.64 | 95.60 | 95.33 |
| E | 96.70 | 98.94 | 93.77 | 97.47 | 93.34 | 97.91 | 94.60 | 98.11 |
| F | 98.99 | 99.37 | 98.83 | 99.20 | 99.76 | 99.52 | 99.19 | 99.36 |

The `OST_BEST`$_3$ model achieved 97.38% $mAP_{50}$ on the cropped images, while the `OST_BEST`$_1$ model had 97.04% $mAP_{50}$ on the original images. These results are significantly better than the results with zero-shot transfer. The precision-recall curves are shown in Fig. 16, and the $F_1$ scores are shown in Fig. 17. The mAP scores are close to optimal and there is only an insignificant deviation between the different training sessions. The $F_1$ score is also relatively high and flat in all cases, indicating that the models are not sensitive to different thresholds.

The performance on the different types of test images is presented in Table X. In general, the models work well, above 94% $mAP_{50}$ in each case. The crowded scenes (group D) have slightly worse performance on average, but the difference is not significant.

Furthermore, the mAP scores of the different classes are presented in Table XI and shown in Fig. 18. All classes perform well, the worst-performing class with the original images being the L-bracket with 92.62% $mAP_{50}$. The precision-recall curves of the `OST_BEST`$_3$ for the different classes are depicted in Fig. 19. Compared to the zero-shot approach, the curves are shifted toward the top-right corner which demonstrates better performance. The proposed confusion matrix of the `OST_BEST`$_3$ with the threshold set to 0.8 is shown in Fig. 20. Almost all the values are in the diagonal, meaning that they are good predictions. However, some instances of L-bracket, U-bracket, and cable shoe were not found by the model. The number of false negative examples can be decreased by lowering the threshold at the expense of possible false positive examples.

TABLE XI
$\text{mAP}_{50}$ Scores of `OST_BEST` Models for the Different Classes

| ID | Train #1 | | Train #2 | | Train #3 | | AVG. | |
|---|---|---|---|---|---|---|---|---|
| | Orig. | Crop. | Orig. | Crop. | Orig. | Crop. | Orig. | Crop. |
| 1 | 93.70 | 93.94 | 90.24 | 92.70 | 93.92 | 94.92 | 92.62 | 93.85 |
| 2 | 94.75 | 92.45 | 92.94 | 93.21 | 95.28 | 94.15 | 94.32 | 93.27 |
| 3 | 93.19 | 94.85 | 92.95 | 93.25 | 92.61 | 94.22 | 92.92 | 94.11 |
| 4 | 96.76 | 96.93 | 96.74 | 97.11 | 96.65 | 96.70 | 96.72 | 96.91 |
| 5 | 99.34 | 99.54 | 99.19 | 99.84 | 98.38 | 99.88 | 98.97 | 99.75 |
| 6 | 99.51 | 99.57 | 99.17 | 99.71 | 99.40 | 99.53 | 99.36 | 99.60 |
| 7 | 98.30 | 98.81 | 98.51 | 98.81 | 97.69 | 98.77 | 98.17 | 98.80 |
| 8 | 98.50 | 98.40 | 96.94 | 98.00 | 98.23 | 98.45 | 97.89 | 98.28 |
| 9 | 99.52 | 98.43 | 97.74 | 98.47 | 99.96 | 99.53 | 99.07 | 98.81 |
| 10 | 96.83 | 97.73 | 93.70 | 96.26 | 96.73 | 97.60 | 95.75 | 97.20 |

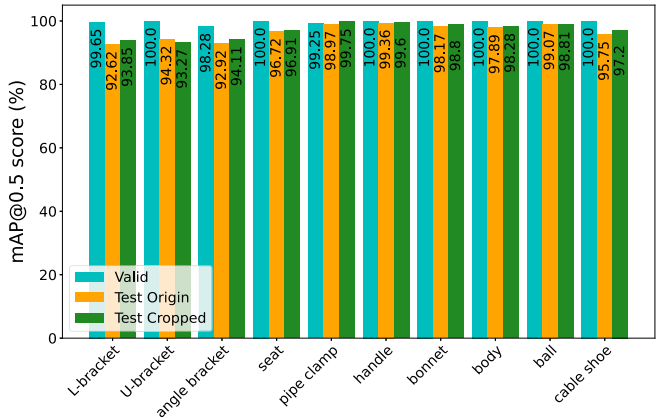

Fig. 18. Average $\text{mAP}_{50}$ scores of the `OST_BEST` models in the different classes.

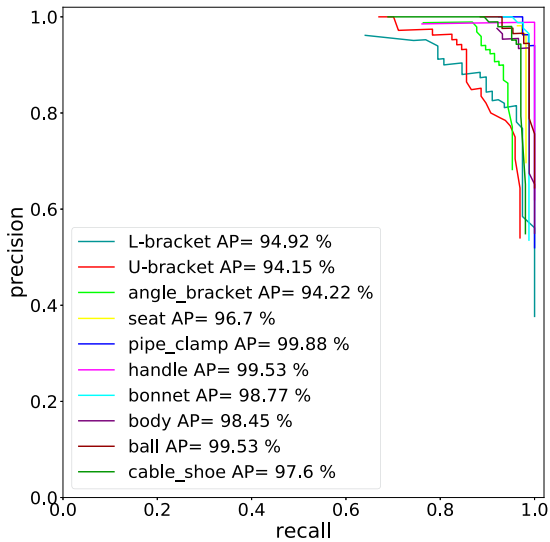

Fig. 19. Class precision-recall curves of the `OST_BEST`$_3$ model on the cropped images.

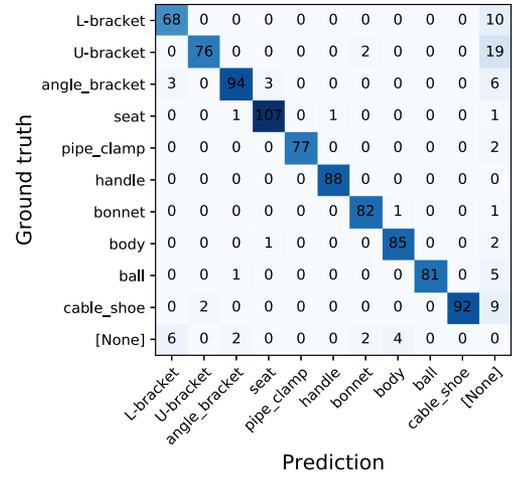

Fig. 20. Confusion matrix of the `OST_BEST`$_3$ model on the cropped images with the threshold set to 0.8.

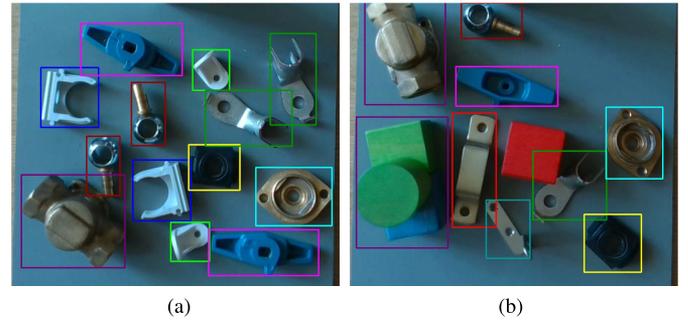

Fig. 21. Accurate (a) and an inaccurate (b) prediction of the `OST_BEST`$_3$ model with the threshold set to 0.8. The color-coding follows Fig. 8. (a) Accurate example. (b) Inaccurate example.

For qualitative evaluation, Fig. 21 shows an accurate prediction and an inaccurate solution. In the latter case, the distractor objects resembling a body object in main characteristics could mislead the model, implying that the model learned an overly general representation of the object. As the quantitative results show, the vast majority of examples is accurate.

## VIII. Ablation Study

In this section, the ablation study of the method is presented, focusing on the different elements of the domain randomization methods, the training data size, and the object detection model.

### A. Seed

In general, the initial random seed of stochastic algorithms can significantly influence their performance. This phenomenon is unpleasant as it makes the algorithms unpredictable. We aim to

TABLE XII
mAP$_{50}$ Scores of ZST_BEST Models With Different Seeds

| | Train #1 | | Train #2 | | Train #3 | | AVG. | |
|---|---|---|---|---|---|---|---|---|
| Seed | Orig. | Crop. | Orig. | Crop. | Orig. | Crop. | Orig. | Crop. |
| 80725 | 79.71 | 78.67 | 81.93 | 84.55 | 78.30 | 80.69 | 79.98 | 81.30 |
| 80725 | 82.29 | 78.31 | 80.99 | 84.31 | 79.60 | 79.23 | 80.96 | 80.62 |
| 53418 | 83.57 | 80.69 | 78.13 | 80.95 | 74.74 | 76.92 | 78.81 | 79.52 |
| 16505 | 85.38 | 86.32 | 81.31 | 82.01 | 82.79 | 81.29 | 83.16 | 83.21 |

TABLE XIII
mAP$_{50}$ Scores of Different ZST Models. -T: Without Texture, -PP: Without Postprocessing, -TPP: Without Postprocessing and Texture

| | Train #1 | | Train #2 | | Train #3 | | AVG. | |
|---|---|---|---|---|---|---|---|---|
| Desc. | Orig. | Crop. | Orig. | Crop. | Orig. | Crop. | Orig. | Crop. |
| BEST | 85.38 | 86.32 | 81.31 | 82.01 | 82.79 | 81.29 | 83.16 | 83.21 |
| -T | 62.83 | 73.26 | 61.08 | 74.07 | 66.59 | 76.80 | 63.50 | 74.71 |
| -PP | 61.40 | 63.14 | 50.58 | 60.62 | 55.63 | 57.51 | 55.87 | 60.42 |
| -TPP | 10.89 | 9.48 | 12.51 | 13.75 | 9.08 | 18.21 | 10.83 | 13.81 |

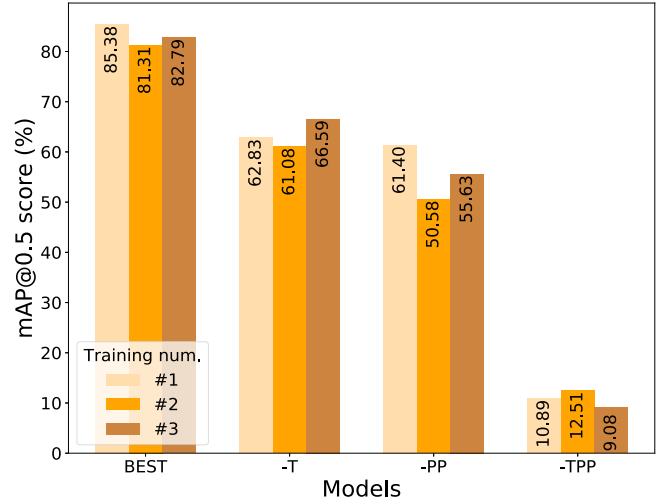

Fig. 22. Results of the ablation study on the original images (ZST models). Model without added textures (-T), without postprocessing methods (-PP), and without both (-TPP).

TABLE XIV
mAP$_{50}$ Scores of ZST_BEST Models With Different Data Sizes

| | Train #1 | | Train #2 | | Train #3 | | AVG. | |
|---|---|---|---|---|---|---|---|---|
| Size | Orig. | Crop. | Orig. | Crop. | Orig. | Crop. | Orig. | Crop. |
| 1000 | 70.07 | 72.12 | 77.94 | 76.60 | 75.37 | 78.12 | 74.46 | 75.61 |
| 4000 | 85.38 | 86.32 | 81.31 | 82.01 | 82.79 | 81.29 | 83.16 | 83.21 |
| 8000 | 84.56 | 85.64 | 86.95 | 83.68 | 81.96 | 81.46 | 84.49 | 83.59 |

measure the influence of the seed of our domain randomization method in the case of the ZST_BEST models. It is important to note that we do not use the same random seed for the training and for the domain randomization method. Table XII shows different seeds (with two equal seeds for reference), with 3 independent training sessions each. In these experiments, we showed that the magnitude of the deviation of the results due to the stochastic training process of the neural network and due to the different seeds of the randomized data generation methods are comparable. Thus, our randomized data generation method is not less robust to a given seed than the stochastic training method itself.

### B. Texture and Postprocessing

Two important factors in our domain randomization method are the random textures of the objects and the postprocessing method. We have generated datasets without these factors. The results are summarized in Table XIII and the results on the original images are shown in Fig. 22. Both the added texture and the postprocessing methods contribute significantly to the performance. Without the added texture, the performance drops to 63.50% and 74.71% mAP$_{50}$ in the case of the original and the cropped images. Without postprocessing, the performance is only 55.87% and 60.42% mAP$_{50}$, respectively. Finally, the performance decreases drastically achieving 10.83% and 13.81% mAP$_{50}$ without the two methods. These experiments show how essential these types of domain randomization methods are. As the average performance of the model on the validation dataset is 99.84% mAP$_{50}$, according to (2), the reality gap shrinks, in case of the original images, on average from 89.01% (-TPP) to 16.68% mAP$_{50}$ (BEST).

### C. Data Size

The size of the training dataset is a key attribute of any machine learning problem. In general, the more data are used in the training, the better its distribution will match the real probability distribution. Nevertheless, this phenomenon does not necessarily apply to the case of knowledge transfer. The results of the performance of the ZST_BEST models for different training data sizes are presented in Table XIV. It is important to note that for the case of 8000 images, the training time was doubled from 5000 to 10 000 iterations. Even though increasing the training data size from 1000 to 4000 allows the model to gain notable performance, doubling the data size from 4000 to 8000 only causes marginal improvement.

### D. Gravity, Positional Disturbance, and Bounding-Box Calculation

In this part of the ablation study, the effect of simulated gravity, the effect of random disturbance around the grid positions, and the effect of replacing the all-point bounding box calculation with the 8-point bounding box calculation (described in Section IV-B4) are measured. The findings are summarized in Table XV. All of the aforementioned factors have a relevant effect on the performance. In the case of cropped images, on average, gravity brings 11.01% mAP$_{50}$, the randomness of object positions contributes 14.22% mAP$_{50}$, and the tight all-point bounding box calculation method is responsible for

TABLE XV

mAP$_{50}$ Scores of ZST Models Without Different Factors. −G: No Gravity, −R: No Randomness in Grid Positions and No Gravity, 8P: 8-Point Bounding Box Calculation

| | Train #1 | | Train #2 | | Train #3 | | AVG. | |
|---|---|---|---|---|---|---|---|---|
| ID | Orig. | Crop. | Orig. | Crop. | Orig. | Crop. | Orig. | Crop. |
| BEST | 85.38 | 86.32 | 81.31 | 82.01 | 82.79 | 81.29 | 83.16 | 83.21 |
| −G | 61.12 | 71.47 | 65.86 | 69.89 | 68.76 | 75.25 | 65.25 | 72.20 |
| −R | 49.12 | 63.07 | 62.00 | 74.38 | 48.18 | 69.53 | 53.10 | 68.99 |
| 8P | 31.09 | 40.66 | 28.98 | 37.77 | 39.19 | 45.93 | 33.09 | 41.45 |

TABLE XVI

mAP$_{50}$ Scores of ZST Models With Different Cutouts At the Postprocessing Methods

| | Train #1 | | Train #2 | | Train #3 | | AVG. | |
|---|---|---|---|---|---|---|---|---|
| ID | Orig. | Crop. | Orig. | Crop. | Orig. | Crop. | Orig. | Crop. |
| BEST | 85.38 | 86.32 | 81.31 | 82.01 | 82.79 | 81.29 | 83.16 | 83.21 |
| CUT1 | 68.24 | 77.18 | 66.93 | 76.77 | 71.85 | 83.53 | 69.01 | 79.16 |
| CUT2 | 82.12 | 78.97 | 78.56 | 83.33 | 72.99 | 69.79 | 77.89 | 77.36 |
| CUT3 | 58.25 | 75.13 | 56.39 | 74.45 | 58.97 | 76.46 | 57.87 | 75.35 |
| CUT4 | 76.00 | 69.76 | 70.17 | 78.62 | 68.26 | 73.60 | 71.48 | 73.99 |
| CUT5 | 70.32 | 74.74 | 75.50 | 78.74 | 74.59 | 74.91 | 73.47 | 76.13 |
| CUT6 | 78.53 | 76.86 | 74.83 | 79.77 | 79.70 | 82.27 | 77.69 | 79.63 |
| CUT7 | 82.25 | 80.26 | 70.22 | 69.95 | 74.51 | 77.48 | 75.66 | 75.90 |
| CUT8 | 75.64 | 75.63 | 75.58 | 80.46 | 80.96 | 84.06 | 77.39 | 80.05 |

a 41.76% mAP$_{50}$ performance gain. In the case of bounding box calculation, the performance drops with the less tight BBs, implying two reasons. First, the ground-truth BBs are tight, thus computing the IOU$_{50}$ with less tight BBs may result in many discarded matches. Second, in the crowded images, the BBs are too extensive, thus, they could significantly overlap each other which may confuse the model.

### E. Cutouts

Some additional experiments were conducted with different types of cutouts at the postprocessing phase of data generation presented in Table XVI. In these experiments, four types of cutouts were considered: rectangles, partly transparent rectangles, circles, and lines. The number of cutouts and the bounds of the randomized sizes of the cutouts varied over the experiments. The results show that none of the cases could achieve a better performance than the ZST_BEST model which does not have this type of domain randomization. Nevertheless, a more thorough evaluation of the effect of different cutouts can be subject to further research.

### F. Faster R-CNN

To test the performance of the data generation process in the case of a two-stage object detection method, we trained the R101-FPN version of the faster R-CNN [6] model using the Detectron2 [66] framework and Pytorch [67]. This model uses the ResNet-101 [61] model with the feature pyramid network [68] backbone. The results are shown in Table XVII. Even though the performance of the model falls behind YOLOv4, it could

TABLE XVII

mAP$_{50}$ Scores of R101-FPN Faster R-CNN Model

| | Zero-shot transfer | | One-shot transfer | |
|---|---|---|---|---|
| | valid | test | valid | test |
| Train #1 | 85.90 | 49.78 | 97.18 | 71.97 |
| Train #2 | 85.78 | 66.73 | 96.495 | 67.70 |
| Train #3 | 86.38 | 55.41 | 96.78 | 70.73 |
| AVG. | 86.02 | 57.31 | 96.82 | 70.13 |

be increased with a more exhaustive hyperparameter search. Moreover, the Darknet framework uses extra data augmentation for training which we did not reproduce for the Detectron2 framework. It is important to note that here, too, considerable performance improvement is achieved by having one real image for training.

The training of the faster R-CNN model was approximately four times faster than in the case of YOLOv4. On the other hand, inference time was around ten times slower with 2 FPS. In conclusion, YOLOv4 outperformed the faster R-CNN approach in performance and in inference time which are the two most relevant factors.

## IX. ROBOTIC APPLICATION

Object detection can be utilized in many ways. Examples are robotic grasping or pick-and-place applications where the robot needs to detect different workpieces and grasp them or move them to specific locations.

In this section, a real-world robotic implementation of our method is presented. The application can serve as proof of concept built upon our previous work [69], where we proposed a 5 C model-based [70] system architecture for visual-servo-guided cyber-physical robotic assembly cells. Relying on the object detection model, the parameters of grasping (micro plan) can be computed.

The robotic system consists of a six-DoF collaborative robot arm equipped with a digital depth camera, a force sensor, and a two-finger gripper. The task of the robot is to detect scattered workpieces (center points and bounding box information), as well as predict grasping poses. The sensors and actuators of the robot and the sim2real computer vision module are connected in a robot control framework based on ROS [71]. The setup is depicted in Fig. 23, while the software components of the robot control framework are shown in Fig. 24

For robotic applications, every component of the system must work reliably an in real time. In order to evaluate the sim2real computer vision module in a new case, three new industrial parts were used, as depicted in Fig. 23. The data generation and training process went without any problems. Thus, within 13 h, the new model was ready to use. As a qualitative evaluation, the robot was programmed to follow a path over the workpieces while streaming the camera data. On a GeForce RTX 3060 GPU, our computer vision model ran with 20 FPS and constantly localized and classified the objects perfectly with more than 98% confidence most of the time, even in significantly different illumination settings and in the presence of distractor objects.

For grasping the workpieces, the grasping pose needs to be estimated and transformed to the robot coordinate system. For estimating the orientation we used principal component analysis on the detected and cropped bounding boxes of the objects, and a standard camera calibration method was applied for the transformation.

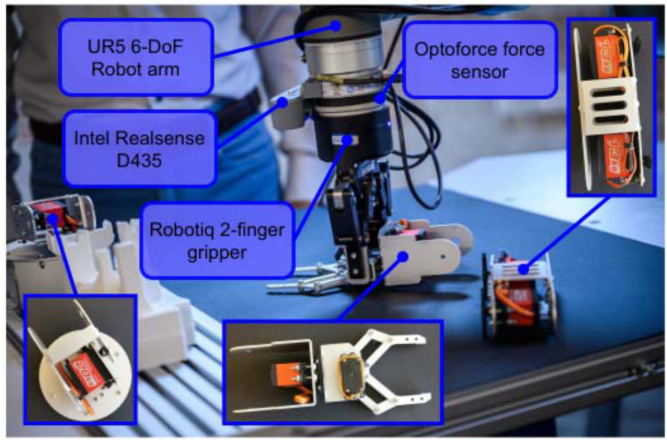

Fig. 23. Setup of the robotic application.

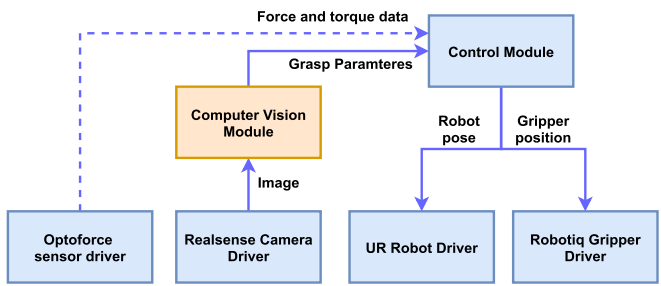

Fig. 24. Information flow in the robotic application. The computer vision module, which is the main topic of this article, is highlighted in orange. In this application, the force sensor was not used (marked with the dashed line). The following software resources were used: [72], [73], [74], [75], [76].

This use case was presented in an exhibition[7] and the implementation of the ROS-based robot control system is available at[8]:

## X. Conclusion

The article presented our sim2real domain randomization method for object detection. As our general aim is to facilitate the trend of new-generation intelligent manufacturing with adaptive robotic applications, our solution needed to be capable of differentiating similar classes using only one real example for training and working in real time. According to our best knowledge, this was the first work thus far that mutually satisfied these constraints in this domain.

As recent works on transfer learning did not concentrate on validation for objects similar to each other, we addressed this phenomenon by creating a dataset with 190 real annotated images of 920 objects of ten classes of industrial workpieces. The dataset was publicly available and could serve as a benchmark for industrial object detection models.

We introduced a novel type of confusion matrix tailored to object detection. It had proven to be useful for finding the root cause of performance loss.

The results presented in the article validated the strengths of our approach. We achieved $86.32\%$ $\text{mAP}_{50}$ in the case of zero-shot transfer, while with one-shot transfer, the best model scored $97.38\%$ $\text{mAP}_{50}$ on the test set. With these experiments, we also demonstrated how to diminish the performance loss caused by similar classes by introducing only one image from the target domain. Even though it was hard to compare solutions to different problems, we believed that these results were better than the ones in the literature considering the complexity of the problem and the size of the synthetic and real training datasets.

In a thorough ablation study, we showed that adding random texture and our postprocessing domain randomization methods were crucial parts of the process. We also found that simulating gravity, random initial placement, and the all-point bounding box calculating method contributed significantly to the performance.

As a proof of concept, we showed that our model works reliably and in real time in a robotic pick-and-place application.

Both the sim2real data generation and training module, and the robot control framework could be used as a freely available, out-of-the-box solution to industrial problems.

In the future, we aim to further improve the performance of the zero-shot learning method. We are also interested in working with point cloud data, exploring the field of adversarial training, and extending our method to be able to predict end-to-end grasping poses as well.

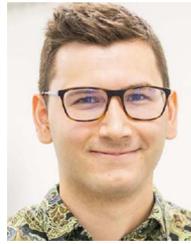
**Dániel Horváth** received the M.Sc. degree in mechatronics from the Budapest University of Technology and Economics, Budapest, Hungary, in 2019, spending one semester each with the Technical University of Denmark, Copenhagen, Denmark, and with the Otto von Guericke University, Magdeburg, Germany. He is currently working toward the Ph.D. degree in computer science with the Eötvös Loránd University, Budapest, Hungary, in collaboration with the Research Laboratory on Engineering and Management Intelligence, Institute for Computer Science and Control, Budapest, Hungary, under the supervision of Gábor Erdős, Zoltán Istenes, and Tomáš Horváth.

His main research interests include computer vision, transfer learning, robotic grasping, and mobile robots.

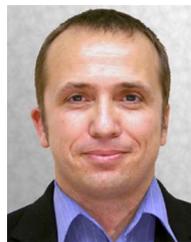
**Gábor Erdős** received the M.Sc. degree in mechanical engineering from the State University of New York at Buffalo, Buffalo, NY, USA, in 1995, and the Ph.D. degree in robitics from the Budapest University of Technology and Economics, Budapest, Hungary, in 2000.

He was a Postdoctoral Researcher with the École Polytechnique Fédérale de Lausanne, Lausanne, Switzerland until 2002. He joined the Institute for Computer Science and Control, Budapest, Hungary, as a Researcher in 2003 and since 2018, he has been the Deputy Head of the Research Laboratory on Engineering and Management Intelligence. His main research interests include robotics, point-cloud and 3-D modeling, digital twin models, multibody kinematics, and virtual manufacturing.

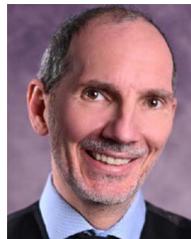
**Zoltán Istenes** received the Ph.D. degree in informatics from the University of Nantes, Nantes, France, in 1997.

Since 1997, he has been an Associate Professor with the Faculty of Informatics, Eötvös Loránd University (ELTE), Budapest, Hungary. Between 2008 and 2020, he was the Manager of the Budapest EIT Digital Doctoral School. He is the Founder of the ELTE Faculty of Informatics Robotics Lab. His main teaching and research interests include computer architectures, artificial intelligence, expert systems, robotics, IoT, UAVs, formal methods, and recently autonomous self-driving vehicles.

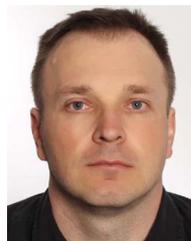
**Tomáš Horváth** received the Ph.D. degree in informatics from the Pavol Jozef Šafárik University, Košice, Slovak Republic, in 2008.

Since 2008, he has been a Faculty Member with Pavol Jozef Šafárik University. Between 2009 and 2012, he was a Postdoctoral Researcher with the University of Hildesheim, Hildesheim, Germany, and between 2015 and 2016, with the University of São Paulo, São Carlos, Brazil. Since 2016, he has been the Head of the Data Science and Engineering Department, Faculty of Informatics, Eötvös Loránd University, Budapest, Hungary. His research interests include relational learning, pattern mining, recommender systems, meta-learning, and automated machine learning.

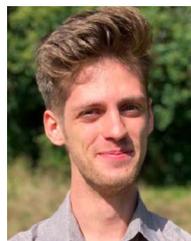
**Sándor Földi** received the B.Sc. degree in mechatronics from the Budapest University of Technology and Economics, Budapest, Hungary, in 2021. He is currently working toward the M.Sc. degree in mathematical modeling and computation with the Technical University of Denmark (DTU), Lyngby, Denmark.

In 2020, he joined the Research Laboratory on Engineering and Management Intelligence, Institute for Computer Science and Control, Budapest, Hungary, to work on sim2real transfer learning with Dániel Horváth and Gábor Erdős. His main research interests include computer vision, transfer learning, and generative adversarial networks.